\pdfoutput=1

\documentclass[11pt]{article}

\usepackage[]{ACL2023}

\usepackage{times}
\usepackage{latexsym}

\usepackage[T1]{fontenc}

\usepackage[utf8]{inputenc}

\usepackage{microtype}
\usepackage{adjustbox}
\usepackage{graphicx}
\usepackage{booktabs}
\usepackage{multirow}
\usepackage{adjustbox}
\usepackage{caption}
\usepackage{xspace}

\usepackage{listings}
\usepackage{amsfonts}
\usepackage{amsmath}
\usepackage{colortbl}
\usepackage{array}
\usepackage{ifthen}

\RequirePackage{type1cm}
\RequirePackage{color}
\RequirePackage{soul}
\setstcolor{blue}
\definecolor{violet}{rgb}{0.5,0.0,0.5}
\newsavebox\bscombox
\newcommand{\bscom}[3][]{%
	\sbox{\bscombox}{\fontsize{8}{9}\selectfont#1#2#3}
	\noindent
	\st{#2}{\selectfont
		\color{blue}#3\ifx\\#1\\\else{\fontsize{8}{9}\selectfont\color{violet}[#1]}\fi
	}
}

\usepackage[raster, skins]{tcolorbox}
\usepackage{graphicx}
\usepackage{subcaption}

\definecolor{mysteryRed}{RGB}{220, 96, 80}
\definecolor{candidateGray}{RGB}{211, 211, 211}
\definecolor{candidateGreen}{RGB}{115, 179, 107}

\usepackage[table,xcdraw]{xcolor} 
\usepackage{inconsolata}

\usepackage{todonotes}

\definecolor{peter}{rgb}{1.0, 0.6, 0.4}
\definecolor{owen}{rgb}{0.55, 0.71, 0.0}

\title{IXAM: Interactive Explainability for Authorship Attribution Models}

\author{
	\textbf{Milad Alshomary \textsuperscript{\dag}},
	\textbf{Anisha Bhatnagar \textsuperscript{\S}},
	\textbf{Peter Zeng\textsuperscript{\ddag}},
	\\
    \textbf{Smaranda Muresan\textsuperscript{\dag}},
    \textbf{Owen Rambow\textsuperscript{\ddag}},
	\textbf{Kathleen McKeown\textsuperscript{\dag}}
	\\
	\\
	\textsuperscript{\dag}Columbia University, New York, USA
	\\
	\textsuperscript{\ddag}Stony Brook University, New York, USA
	\\
    \textsuperscript{\S}University of Pennsylvania, Philadelphia, USA
    \\
	\small{
		\textbf{Correspondence:} \href{ma4608@columbia.edu}{ma4608@columbia.edu}
	}
}

\begin{document}
\maketitle

\begin{abstract}
We present IXAM, an {\bf I}nteractive e{\bf X}plainability framework for {\bf A}uthorship Attribution {\bf M}odels. Given an authorship attribution (AA) task and an embedding-based AA model, our tool enables users to interactively explore the model's embedding space and construct an explanation of the model's prediction as a set of writing style features at different levels of granularity. Through a user evaluation, we demonstrate the value of our framework compared to predefined stylistic explanations.

\end{abstract}

\section{Introduction}
\label{sec:intro}

Authorship attribution (AA) is the task of identifying the author of a given text. It involves an automatic analysis and comparison of writing style features of different documents. The task is particularly important for domains such as forensic linguistics, where an investigator might use evidence from texts as part of their testimony in criminal trials \cite{tiersma2002linguist}. State-of-the-art models for this task are embedding-based, where documents are matched based on their vector similarities in a learned latent space \cite{rivera-soto-etal-2021-learning, wegmann-etal-2022-author}. Therefore, these models are inherently black boxes and thus the ability to explain their predictions is entirely lacking. Despite the large body of research on explainability, very little work has been done to explain the latent space of AA models. Most of this work is exploratory, aiming to determine whether AA models capture specific style features \cite{lyu2023representation, wegmann-etal-2022-author} in their latent space. Recently, \citet{alshomary2025latent} proposed an approach to automatically construct an interpretable space from the AA model's latent space and utilize it to explain the model's prediction. 

\begin{figure*}
    \centering
    \includegraphics[width=0.9\textwidth]{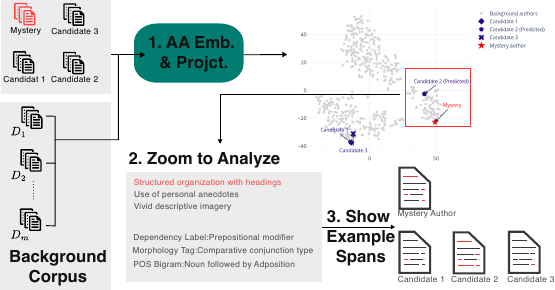}
    \caption{Our approach for building an interactive tool for exploring style explanations for AA models.}
    \label{fig:main-approach}
\end{figure*}

In this paper, we introduce IXAM\footnote{Link to the Tool: \url{https://huggingface.co/ExplainabiliyForAATeam}}, an open-source tool that enables users to construct interactive explanations of the AA model's predictions. Unlike the work of \cite{alshomary2025latent} that provides static explanations of the prediction, our system, through an interactive interface, allows users to explore different regions of the latent space and inspect their common writing style features, leading to more user involvement and richer explanations of the model's prediction. As shown in Figure \ref{fig:main-approach}, users of our system can construct post-hoc explanations of the model's prediction through (1) visualizing a two-dimensional approximation of the latent space, showing the texts in question with respect to a background corpus, (2) zooming into a specific cluster of authors to be analyzed by the system in terms of their common writing style, and (3) highlighting the analyzed writing style features in the given texts. Our system's explanations rely on analyzing the common writing style of clusters of authors in the latent space through utilizing LLMs and the Gram2Vec framework \cite{zeng2025gram2vec}. While the former provides context-relevant high-level style analysis of texts covering a wide range of feature types, the latter generates linguistic features directly traceable to the documents.

We design a user study to compare the user experience of our tool with the explainability framework of \citet{alshomary2025latent}. The evaluation involves providing users with an AA task consisting of a mystery author and three candidate authors, along with the model's prediction and explanations. Users must determine whether the model's prediction is correct, provide supporting explanations, and indicate their level of confidence in their decision. We conduct a small user study with three participants, each completing ten tasks (five for each system). After working on these tasks, participants completed a survey comparing their experience with the new tool versus the baseline framework. The results show that all three users highly value the features of our tools compared to the baseline. Further analysis demonstrated that the explanations provided by our tool had a higher impact on users' confidence scores compared to the baseline.

\section{Related Work}
SOTA authorship attribution systems are embedding-based: pre-trained transformer models that are fine-tuned on large corpora to learn style representations. \citet{rivera-soto-etal-2021-learning} pool representations from multiple text spans and apply a contrastive objective to bring same-author embeddings closer together, while \citet{wegmann-etal-2022-author} explicitly encourages the embedding space to encode stylistic rather than topical features. Despite their strong performance, these approaches rely on latent representations that offer limited explainability. Existing efforts to interpret such embeddings (e.g., \citealp{wegmann-etal-2022-author}) do not evaluate the utility of these interpretations for end-user explanations. Another approach augments inherently interpretable systems by fine-tuning a transformer model to correct the interpretable system \citep{zeng-etal-2025-residualized}, but there is no immediate explanation of the correction. \cite{patel-etal-2023-learning} attempts to learn interpretable style embeddings by prompting LLMs.

To our knowledge, \citet{alshomary2025latent} are among the few who generate explanations for model predictions, but their method depends on predefined clusters in the latent space. We emphasize that while there are some other methods for explaining authorship attribution, there have been no systems created to demonstrate their efficacy when provided to real users. 

\section{System Description}
In the following, we provide an overview of the task of authorship attribution (AA) and embedding-based AA models for this task. We then describe the main components of our proposed system. Finally, we describe how the system performs on a sample case study.

\subsection{Authorship Attribution}

Authorship attribution is the task of determining the author of a given piece of writing based on stylistic and linguistic features extracted from the text. In this paper, as shown in Figure~\ref{fig:authorship-task}, we adopt the following task format: given a mystery text and three candidate authors, select one of the candidates to be the author of the mystery text. As mentioned, embedding-based AA models learn from large corpora a function that maps documents into a latent embedding space and then use this function to predict the authors of new documents. Therefore, explaining a model prediction involves (1) identifying regions of interest in the embedding space, and (2) describing their shared authorial style.

\subsection{Interactive Explainability System}

As highlighted in Figure \ref{fig:main-approach}, our system takes as input (1) a background corpus to be used as context for identifying specific regions in the latent space, (2) an AA task defined by the mystery author and three candidate authors (task authors), and (3) any embedding-based AA model.
Through interactive exploration, users can visualize a 2D projection of the model's latent space, zoom into a specific cluster of authors, and highlight common writing style features of these clusters in the task authors' texts. Instead of a predefined single explanation, this interactivity allows users to construct a wider set of different hypotheses that explain the model's prediction. In the following, we present relevant implementation details.

\paragraph{1. Latent Space Visualization.} Our system first runs the background corpus and the task authors' documents through the AA model to obtain their latent embeddings. It then projects the embeddings into a two-dimensional space using the t-SNE algorithm and visualizes the result in a scatter plot. As shown in Figure \ref{fig:main-approach}, this allows the user to inspect the distance between the mystery author, the predicted author, and other candidates, as well as to identify what clusters of background authors these task authors belong to. The user can then zoom and pan to a specific region in the latent space to analyze its common authorial style.

\paragraph{2. Analyzing the authorial style.} When a user requests an analysis of a zoomed-in region in the latent space, our system computes two sets of representative writing style features for it.  First, we obtain dynamically generated using large language models (LLMs). LLMs have been employed for writing style analysis in textual data \cite{alshomary2025latent, patel2023learning}, as they obviate the need for a predefined set of stylistic features and are able to induce context-relevant style features. In our system, we provide \texttt{gpt-4o}\footnote{\url{https://openai.com/index/hello-gpt-4o/}} with texts from up to 10 authors within the selected zoomed-in region and prompt it to characterize the writing style that is shared across all 10 authors.
Second, we use the interpretable Gram2vec approach of \citet{zeng2024gram2vec}. Gram2Vec creates an interpretable linguistic representation of an author’s style. It extracts the normalized relative frequencies of grammatical features present in a given text. In our system, we compute the most frequent Gram2vec features in the texts of the authors in the zoomed-in region. 

\paragraph{3. Highlighting style features in Texts.} Upon request, given an LLM-generated or a Gram2Vec-generated style feature, our system highlights all text spans of the task authors' texts that contain this feature. This highlighting allows the user to verify the exact part of the text that was claimed to have the specific feature. We extract these spans from the texts of the task authors using the Gram2Vec library (for the Gram2Vec features) and prompting \texttt{gpt-4o} to find the spans relevant to each of the LLM-generated style features (see second Prompt in \ref{fig:prompts}.

\paragraph{Technical Specifications}
Our tool uses caching at three levels: (1) t-SNE projections of the entire embedding space, (2) author subsets in zoomed areas, and (3) writing style features and related text spans from GPT-4o and Gram2Vec, all scoped to the user's AA task and specific zoomed region if applicable. The interface is a modular web app using the Gradio framework. It is optimized for interactive, per-task, and per-region exploration on standard CPU hardware. Besides predefined AA models and tasks, the tool allows users to work with any HuggingFace-based Sentence Transformer model and upload their own AA tasks.

\subsection{Case Study}
\begin{figure*}[t!]
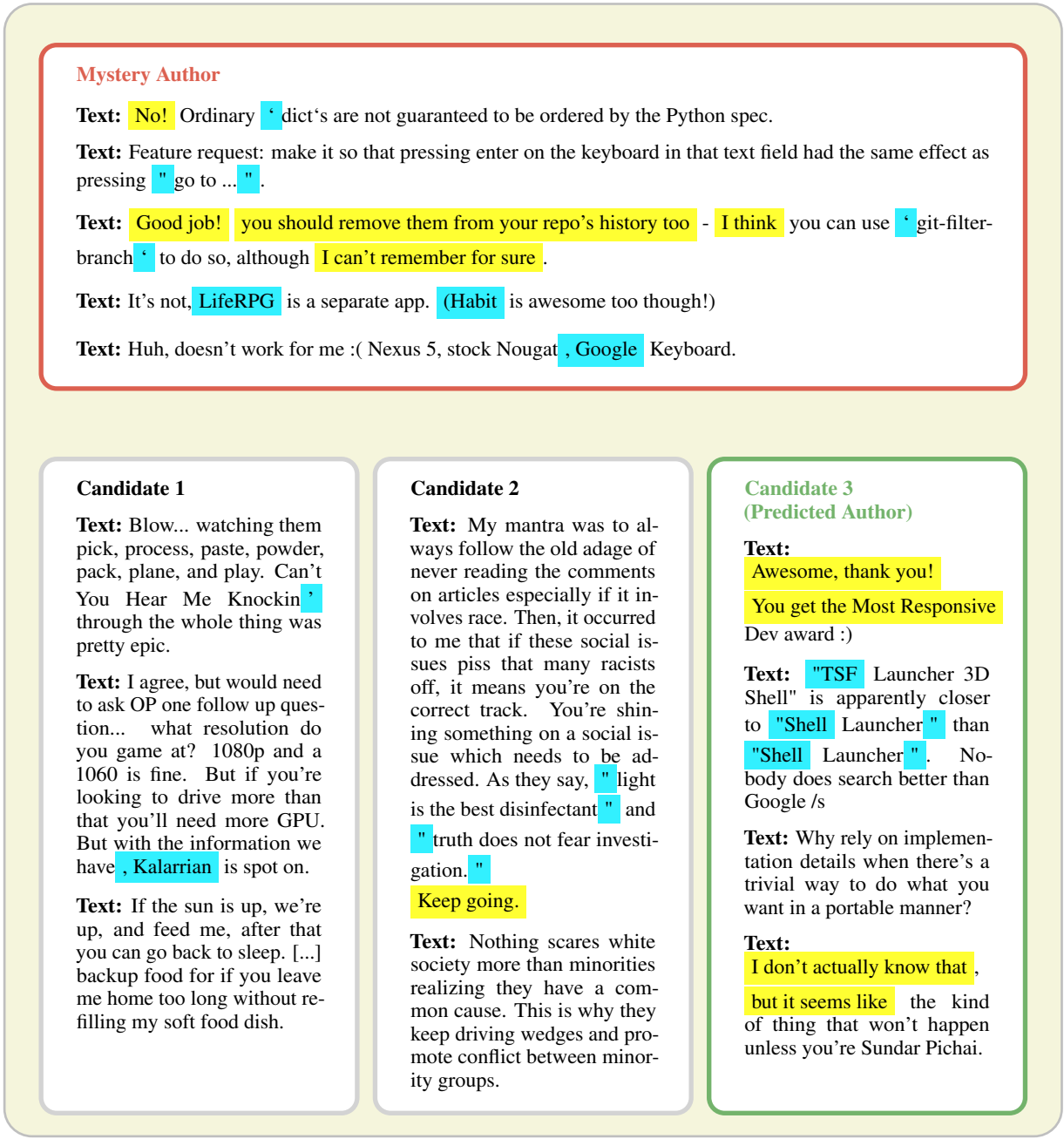

\centering

\definecolor{lightBeige}{RGB}{245, 245, 220} 
\definecolor{highYellow}{RGB}{255, 255, 50} 
\definecolor{highBlue}{RGB}{50, 240, 255}   

\newcommand{\hlyellow}[1]{\colorbox{highYellow}{#1}}
\newcommand{\hlblue}[1]{\colorbox{highBlue}{#1}}

\begin{tcolorbox}[
    colback=lightBeige, 
    colframe=gray!50!white, 
    boxrule=1pt, 
    arc=4mm, 
    enlarge top by=3mm,    
    enlarge bottom by=3mm, 
    enlarge left by=3mm,   
    enlarge right by=3mm,  
    width=\linewidth, 
    nobeforeafter 
]

\tcbset{
    myboxstyle/.style={
        arc=2mm,                
        colback=white,          
        fonttitle=\bfseries,    
        boxrule=2pt,            
        valign=top,             
        attach boxed title to top left={yshift=-4mm, xshift=5mm},
        boxed title style={
            arc=1mm,            
            colback=white,      
            colframe=white      
        }
    }
}

\begin{tcolorbox}[
    myboxstyle,
    title=Mystery Author,
    colframe=mysteryRed, 
    width=\linewidth     
]
\small 
    \textbf{\textcolor{mysteryRed}{Mystery Author}}
    \par\vspace{2mm}
    \textbf{Text:} \hlyellow{No!} Ordinary \hlblue{`}dict`s are not guaranteed to be ordered by the Python spec.
    \par\vspace{2mm} 
    \textbf{Text:} Feature request: make it so that pressing enter on the keyboard in that text field had the same effect as pressing \hlblue{"}go to ...\hlblue{"}.
    \par\vspace{2mm}
    \textbf{Text:} \hlyellow{Good job!} \hlyellow{you should remove them from your repo's history too} - \hlyellow{I think} you can use \hlblue{`}git-filter-branch\hlblue{`} to do so, although \hlyellow{I can't remember for sure}.
    \par\vspace{2mm}
    \textbf{Text:} It's not,\hlblue{LifeRPG} is a separate app. \hlblue{(Habit} is awesome too though!)
    \par\vspace{2mm}
    \textbf{Text:} Huh, doesn't work for me :( Nexus 5, stock Nougat\hlblue{, Google} Keyboard.
\end{tcolorbox}

\vspace{3mm} 

\begin{tcbitemize}[
    myboxstyle,
    raster columns=3,
    raster equal height,
    colframe=candidateGray, 
]

\tcbitem[title=Candidate 1]
    \small
    \textbf{Candidate 1}
    \par\vspace{2mm}
    \textbf{Text:} Blow... watching them pick, process, paste, powder, pack, plane, and play. Can't You Hear Me Knockin\hlblue{'} through the whole thing was pretty epic.
    \par\vspace{2mm}
    \textbf{Text:} I agree, but would need to ask OP one follow up question... what resolution do you game at? 1080p and a 1060 is fine. But if you're looking to drive more than that you'll need more GPU. But with the information we have\hlblue{, Kalarrian} is spot on.
    \par\vspace{2mm}
    \textbf{Text:} If the sun is up, we're up, and feed me, after that you can go back to sleep. [...] backup food for if you leave me home too long without refilling my soft food dish.
    \par\vspace{2mm}

\tcbitem[title=Candidate 2]
    \small
    \textbf{Candidate 2}
    \par\vspace{2mm}
    \textbf{Text:} My mantra was to always follow the old adage of never reading the comments on articles especially if it involves race. Then, it occurred to me that if these social issues piss that many racists off, it means you're on the correct track. You're shining something on a social issue which needs to be addressed. As they say, \hlblue{"}light is the best disinfectant\hlblue{"} and \hlblue{"}truth does not fear investigation.\hlblue{"}
    
    \hlyellow{Keep going.}
    
    \par\vspace{2mm}
    \textbf{Text:} Nothing scares white society more than minorities realizing they have a common cause. This is why they keep driving wedges and promote conflict between minority groups.

\tcbitem[
    title=Candidate 3 (Predicted Author),
    colframe=candidateGreen 
]
    \small
    \textbf{\textcolor{candidateGreen}{Candidate 3}}
    
    \textbf{\textcolor{candidateGreen}{(Predicted Author)}}
    \par\vspace{2mm}
    \textbf{Text:} \hlyellow{Awesome, thank you!} \hlyellow{You get the Most Responsive} Dev award :)
    \par\vspace{2mm}
    \textbf{Text:} \hlblue{"TSF} Launcher 3D Shell" is apparently closer to \hlblue{"Shell} Launcher\hlblue{"} than \hlblue{"Shell} Launcher\hlblue{"}. Nobody does search better than Google /s
    \par\vspace{2mm}
    \textbf{Text:} Why rely on implementation details when there's a trivial way to do what you want in a portable manner?
    \par\vspace{2mm}
    \textbf{Text:} \hlyellow{I don't actually know that}, \hlyellow{but it seems like} the kind of thing that won't happen unless you're Sundar Pichai.
    \par\vspace{2mm}
    
\end{tcbitemize}

\end{tcolorbox}

\caption{An task example consisting of a mystery and three candidate authors. For presentation purpose, we show the highlighting of spans of the top three features of both LLM and Gram2Vec feature types. The top three LLM-derived style features are "Exclamations", "Expressions of uncertainty", and "Direct address". The top three Gram2Vec features are "Part-of-Speech Bigram:Punctuation followed by Proper noun", "Morphology Tag:Quotation punctuation type", and "Morphology Tag:Initial punctuation". It is clear that the Mystery Author's texts and Candidate 3's texts have the most spans highlighted of the different feature types.}
\label{fig:authorship-task}
\end{figure*}

We showcase our tool on an AA case study (Figure \ref{fig:authorship-task}), using the AA model from \citet{wegmann-etal-2022-author} as the black-box predictor to be explained\footnote{A detailed video presentation can be found \href{https://drive.google.com/file/d/1bT8Sf3ooD3PNc2XRagWUASwImVE2Yo9h/view}{here}}. 

\begin{figure*}
    \centering
    \includegraphics[scale=1.0]{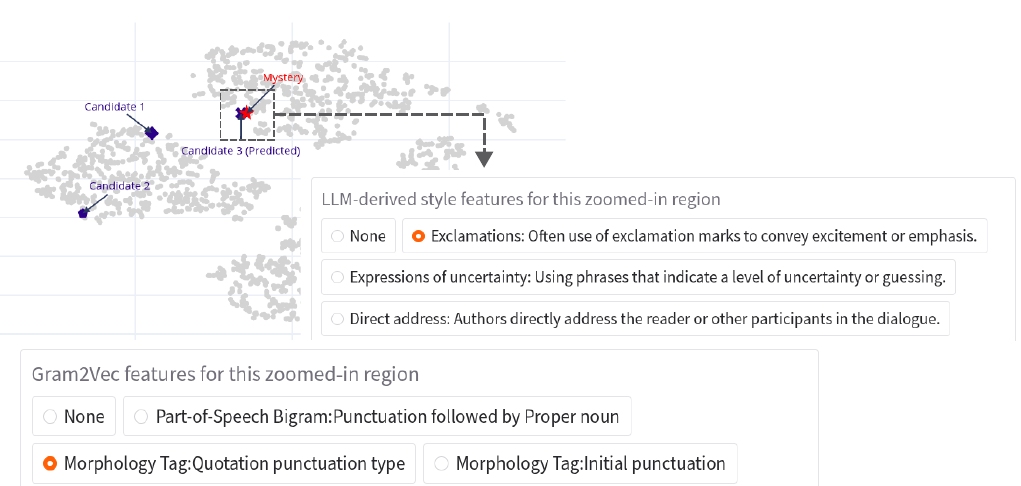}
    \caption{Our tool allows users to visualize the latent space of the AA model (top left), inspect a relevant subregion to see its common writing style features (top right), and show example spans from the text of the task authors for each of the features}
    \label{fig:interface_example}
\end{figure*}

First, as shown in Figure \ref{fig:interface_example}, {\bf the user can explore a 2D projection of the AA model's latent space}, visually navigating clusters of authors and salient regions in the space. The task authors (the mystery author and the three candidates) are highlighted in this projection. The user immediately sees that the mystery author and candidate 3 form a tight cluster, together with several background authors (gray dots in the figure), providing a clear visual rationale for the model’s prediction. In contrast, candidates 1 and 2 appear isolated and scattered elsewhere in the space.

Second, {\bf the user can inspect the characteristic writing style of any region in the latent space} by zooming into it. A particularly relevant region is the neighborhood shared by the mystery author and candidate 3. Zooming into this area, as shown in Figure \ref{fig:interface_example}, surfaces representative style features, grouped by generation method. Here, the user uncovers a distinctive style: frequent exclamations, expressions of uncertainty, direct address to the reader, and heavy use of quotation.

Third, {\bf the user can highlight the exact spans in the task authors’ texts that exhibit each representative style feature}, as illustrated in Figure \ref{fig:authorship-task}. This enables direct visualization of how strongly each text reflects the identified style patterns. Through this interaction, the user can ground the model’s prediction in concrete evidence: the mystery author and candidate 3 share a common writing style defined by exclamation marks, expressions of uncertainty, direct reader address, and extensive quotation.
\section{Evaluation}
Through a user study, we compare the user experience of our interactive explanation tool IXAM with predefined static explanations \cite{alshomary2025latent}. We design an evaluation study where we present users with (1) an authorship attribution task consisting of a mystery author and three candidates, (2) the model's prediction, and (3) the evaluated explanation system. The user's task is then to determine whether the model predicted the correct candidate, provide a supporting explanation, and indicate their level of confidence in the decision. The confidence assessment indicates whether the explanations had a positive impact on the user's decision. Additionally, participants must complete a survey about their experience in interacting with the system, comparing it with the baseline.

\paragraph{Data and AA model} For the user study, we take the AA model proposed by \citet{wegmann-etal-2022-author} as the black-box to be explained. 
We use the Reddit corpus collected by \citet{baumgartner2020pushshift}, which contains posts and comments from Reddit users, as the background corpus and for the AA tasks. To construct the background corpus, we sample 10k authors from the training split with a minimum of 3 texts and a maximum of 10. We apply the clustering approach of \cite{alshomary-etal-2021-counter} to discover salient regions in the latent space and eliminate outlier authors (authors not assigned to any cluster), resulting in 2252 authors. This background corpus, along with its clustering labels, is used in both our explainability system and the baseline. The task authors are randomly sampled from the test split.

\paragraph{Pre-defined Explanation Baseline} To construct the baseline explanations, we follow \cite{alshomary2025latent}, who proposed a framework for explaining embedding-based AA models through identifying salient regions in the latent space by clustering training authors and describing their writing style using LLMs. Unlike IXAM, this approach relies on precomputed clusters of training authors to explain the prediction. In particular, given a training corpus of authors' texts, we generate a latent representation of these authors using the AA model and cluster them to identify salient regions in the latent space. We then prompt \texttt{gpt-4o} to describe the common writing style of this cluster of authors. Finally, to generate style explanations for a given AA task, we compute the latent representation for each task author and identify their closest precomputed cluster. The style explanation for each author is then the style description of its closest cluster. We present sample explanations in Figure \ref{fig:baseline-explanations}.

\begin{table*}[]
    \centering
    \begin{tabular}{lrrrrrr}
         & \multicolumn{3}{c}{\bf Baseline} & \multicolumn{3}{c}{\bf IXAM (Our Tool)} \\
         \cmidrule(l){2-4} \cmidrule(l){5-7}
       \bf User  & \bf Acc. & \bf Conf/Correct & \bf Conf/Wrong & \bf Acc. & \bf Conf/Correct & \bf Conf/Wrong \\
       \midrule
       U1  & 1.0 & 4.7 & 5.0 & 0.8 & 5.0 & 3.0 \\
       U2  & 0.8 & 3.3 & 3.0 & 0.8 & 4.7 & 3.0\\
       U3  & 0.8 & 4.7 & 5.0 & 1.0 & 5.0 & 3.5\\
    \end{tabular}
    \caption{The accuracy of the annotators of identifying if the model predicts correctly (Acc.), and the reported confidence (scale of 1 to 5) on when the model made correct (Conf/Correct) and wrong (Conf/Wrong) predictions.}
    \label{tab:main-results}
\end{table*}

\begin{figure}
    \centering
    \includegraphics[]{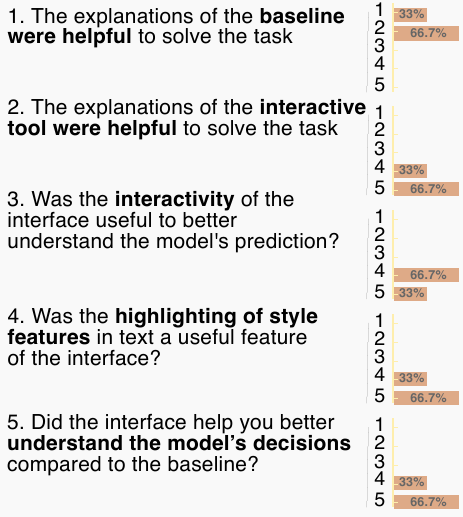}
    \caption{Survey questions along with the scores distributions given by the three participants in the study.}
    \label{fig:survey-questions-1}
\end{figure}

\paragraph{User Study} Since a single task involves reading the texts of four authors and exploring the explanation of the model prediction, it takes up to 10 minutes. Therefore, we conducted a small user study with three participants, each working on 10 AA tasks, five tasks for each evaluated system, of which three were correctly predicted by the model, and two were incorrectly predicted. 
The users in our evaluation were graduate students with experience in natural language processing and familiarity with the authorship attribution task. The user study begins with the participant completing 5 tasks with the aid of predefined explanations presented in Label Studio\footnote{\url{https://labelstud.io/}}. In the second part, one of the authors of the paper demoed our interactive interface to the participant, then the participant was asked to work on the second set of five AA tasks using IXAM. Finally, the participant answered a set of questions as shown in Figure \ref{fig:survey-questions-1} on a 5-point Likert scale.


\paragraph{Results} The answers for the survey questions are in Figure \ref{fig:survey-questions-1}. The three annotators gave scores of 4 or 5 when rating the helpfulness of the explanations provided by our interface, while giving low scores (1 and 2) for the baseline explanations. Moreover, we assessed the agreement among participants regarding which style features were deemed relevant to the model's prediction. For the baseline explanations, there is a 3\% overlap in the selected style features, which is significantly lower than the 38\% overlap observed when using IXAM for explanations. This high level of agreement highlights the plausibility of the explanations IXAM provides.

The results of participants' accuracy in identifying whether the model predicted correctly are not conclusive (Table \ref{tab:main-results}). The accuracy score for U1 decreased when IXAM was used, while it increased for U3 and remained unchanged for U2. 
However, when we look at the users' reported confidence, an interesting distinction emerges.  For the baseline, there was no major distinction in confidence between the cases where the system predictions were correct and when they were not correct.  In contrast, for IXAM, we see a clear decrease in confidence when the system makes a wrong prediction.  This suggests that IXAM's explanations are more readily interpretable by the users than the baseline explanations, since IXAM's explanations for the wrong predictions should in fact make less sense.


\section{Discussion and Conclusion}
In this paper, we present a demo system for analyzing the latent space of AA models to better understand their predictions. The system lets users inspect the latent space via 2D projections, examine stylistic features common to specific subregions, and highlight text spans that exhibit these features. It supports diverse users, including forensic linguists and technical developers, by helping them investigate cases and interpret model behavior. Because existing AA explainability tools are limited, we compare against a simple baseline with precomputed, cluster-based explanations; its static nature is a major drawback relative to our interactive tool. A user study shows participants strongly prefer our system over the baseline.

\section{Limitations}
Our tool provides strongly plausible explanations to the model's prediction, yet due to its post-hoc nature, these explanations might not be faithful to the model's prediction. An extension to our work could consider performing suggested edits on the texts and testing the model's prediction to analyze the causality between these features and the prediction. The 2D projection of the latent space, while a useful approximation of salient regions in the space, is still an estimate that might generate noise. In our experiments, we worked with Reddit data as the background training corpus, which represents a similar domain to the training data of the AA model. However, in scenarios where we don't have the training data of the black box model, there might be a domain shift, and this would maybe lead to misalignment.

\bibliography{anthology,custom}

@misc{zeng2025gram2vec,
      title={Gram2Vec: An Interpretable Document Vectorizer}, 
      author={Peter Zeng and Hannah Stortz and Eric Sclafani and Alina Shabaeva and Maria Elizabeth Garza and Daniel Greeson and Owen Rambow},
      year={2025},
      eprint={2406.12131},
      archivePrefix={arXiv},
      primaryClass={cs.CL},
      url={https://arxiv.org/abs/2406.12131}, 
}

@inproceedings{alshomary2025latent,
  title={Latent Space Interpretation for Stylistic Analysis and Explainable Authorship Attribution},
  author={Alshomary, Milad and Ri, Narutatsu and Apidianaki, Marianna and Patel, Ajay and Muresan, Smaranda and McKeown, Kathleen},
  booktitle={Proceedings of the 31st International Conference on Computational Linguistics},
  pages={1124--1135},
  year={2025}
}

@inproceedings{patel2023learning,
  title={Learning Interpretable Style Embeddings via Prompting LLMs},
  author={Patel, Ajay and Rao, Delip and Kothary, Ansh and Mckeown, Kathleen and Callison-Burch, Chris},
  booktitle={Findings of the Association for Computational Linguistics: EMNLP 2023},
  pages={15270--15290},
  year={2023}
}

@inproceedings{baumgartner2020pushshift,
  title={The pushshift reddit dataset},
  author={Baumgartner, Jason and Zannettou, Savvas and Keegan, Brian and Squire, Megan and Blackburn, Jeremy},
  booktitle={Proceedings of the international AAAI conference on web and social media},
  volume={14},
  pages={830--839},
  year={2020}
}

@article{tiersma2002linguist,
  title={The linguist on the witness stand: forensic linguistics in American courts},
  author={Tiersma, Peter Meijes and Solan, Lawrence},
  journal={Language},
  volume={78},
  number={2},
  pages={221--239},
  year={2002},
  publisher={Linguistic Society of America}
}

@inproceedings{lyu2023representation,
  title={Representation of Lexical Stylistic Features in Language Models’ Embedding Space},
  author={Lyu, Qing and Apidianaki, Marianna and Callison-Burch, Chris},
  booktitle={Proceedings of the 12th Joint Conference on Lexical and Computational Semantics (* SEM 2023)},
  pages={370--387},
  year={2023}
}

@inproceedings{zeng-etal-2025-residualized,
    title = "Residualized Similarity for Faithfully Explainable Authorship Verification",
    author = "Zeng, Peter  and
      Alipoormolabashi, Pegah  and
      Mun, Jihu  and
      Dey, Gourab  and
      Soni, Nikita  and
      Balasubramanian, Niranjan  and
      Rambow, Owen  and
      Schwartz, H.",
    editor = "Christodoulopoulos, Christos  and
      Chakraborty, Tanmoy  and
      Rose, Carolyn  and
      Peng, Violet",
    booktitle = "Findings of the Association for Computational Linguistics: EMNLP 2025",
    month = nov,
    year = "2025",
    address = "Suzhou, China",
    publisher = "Association for Computational Linguistics",
    url = "https://aclanthology.org/2025.findings-emnlp.856/",
    doi = "10.18653/v1/2025.findings-emnlp.856",
    pages = "15824--15837",
    ISBN = "979-8-89176-335-7"
}

@inproceedings{patel-etal-2023-learning,
    title = "Learning Interpretable Style Embeddings via Prompting {LLM}s",
    author = "Patel, Ajay  and
      Rao, Delip  and
      Kothary, Ansh  and
      McKeown, Kathleen  and
      Callison-Burch, Chris",
    editor = "Bouamor, Houda  and
      Pino, Juan  and
      Bali, Kalika",
    booktitle = "Findings of the Association for Computational Linguistics: EMNLP 2023",
    month = dec,
    year = "2023",
    address = "Singapore",
    publisher = "Association for Computational Linguistics",
    url = "https://aclanthology.org/2023.findings-emnlp.1020/",
    doi = "10.18653/v1/2023.findings-emnlp.1020",
    pages = "15270--15290"
}
\bibliographystyle{acl_natbib}

\appendix

\section{Prompts Used in the paper}
The following are the prompts used to distill LLM-based style features.

\begin{figure}[ht]
\begin{minipage}{.45\textwidth}

\begin{lstlisting}[basicstyle=\ttfamily\tiny, frame=single, breaklines=false, caption={Identifying Common Style Features}, label={lst:societal_impact_eval_prompt},xleftmargin=0pt, xrightmargin=0pt]

You are a forensic linguist who knows how to analyze
linguistic and stylometric similarities between texts.

Identify {max_num_feats} writing style features that
are common between the authors' texts.

Author Texts:
{author_texts}
\end{lstlisting}

\begin{lstlisting}[basicstyle=\ttfamily\tiny, frame=single, breaklines=false, caption={Idenitfying example text spans}, label={lst:societal_impact_eval_prompt},xleftmargin=0pt, xrightmargin=0pt]

You are a linguistic specialist. Given a writing sample
and a list of descriptive features, identify the exact
text spans that demonstrate each feature.
    
Important:
- The headers like "Document 1:" etc are NOT part of the
original text - ignore them.
- For each feature, even if there is no match, return an 
empty list.
- Only return exact phrases from the text.
- Use the EXACT feature names as JSON keys - do not 
paraphrase or shorten them.


Respond in this EXACT JSON format 
(use these exact keys, populate the lists with the
extracted text spans):
{feature: [] for feature in features}

Text:
\"\"\"{text}\"\"\"

Style Features:
{features}
\end{lstlisting}

\end{minipage}
\caption{Prompts that are used in our system}
\label{fig:prompts}
\end{figure}

\begin{figure*}[t!]
\centering

\definecolor{lightBeige}{RGB}{245, 245, 220} 
\definecolor{highYellow}{RGB}{255, 255, 50} 
\definecolor{highBlue}{RGB}{50, 240, 255}   

\newcommand{\hlyellow}[1]{\colorbox{highYellow}{#1}}
\newcommand{\hlblue}[1]{\colorbox{highBlue}{#1}}

\begin{tcolorbox}[
    colback=lightBeige, 
    colframe=gray!50!white, 
    boxrule=1pt, 
    arc=4mm, 
    enlarge top by=3mm,    
    enlarge bottom by=3mm, 
    enlarge left by=3mm,   
    enlarge right by=3mm,  
    width=\linewidth, 
    nobeforeafter 
]

\tcbset{
    myboxstyle/.style={
        arc=2mm,                
        colback=white,          
        fonttitle=\bfseries,    
        boxrule=2pt,            
        valign=top,             
        attach boxed title to top left={yshift=-4mm, xshift=5mm},
        boxed title style={
            arc=1mm,            
            colback=white,      
            colframe=white      
        }
    }
}

\begin{tcolorbox}[
    myboxstyle,
    title=Mystery Author,
    colframe=mysteryRed, 
    width=\linewidth     
]
\tiny 
    \textbf{\textcolor{mysteryRed}{Mystery Author}}
    \par\vspace{2mm}

    Text:
    No! Ordinary `dict`s are not guaranteed to be ordered by the Python spec.
    \par
    Text:
    Feature request: make it so that pressing enter on the keyboard in that text field had the same effect as pressing "go to ...".
    \par
    Text:
    Good job! You should remove them from your repo's history too - I think you can use `git-filter-branch` to do so, although I can't remember for sure.
    \par
    Text:
    It's not, LifeRPG is a separate app. (Habit is awesome too though!)
    \par
    Text:
    Huh, doesn't work for me :( Nexus 5, stock Nougat, Google Keyboard.
    \par\vspace{2mm}
    {\bf Explanation of the style}\par
    The authors' writing styles are distinguished by a blend of informal language and vibrant humor, forming an accessible and entertaining narrative voice. [...]. They enrich their narrative by interweaving intertextual references, drawing from cultural touchstones and external media to provide context and depth. This layered storytelling is supported by a diverse sentence structure, balancing brevity with detail to sustain interest and dynamism. [...]. Collectively, these stylistic features create a narrative voice that is both engaging and distinctively lively, inviting readers into a dialogue rather than a monologue.
\end{tcolorbox}

\vspace{3mm} 

\begin{tcbitemize}[
    myboxstyle,
    raster columns=3,
    raster equal height,
    colframe=candidateGray, 
]

\tcbitem[title=Candidate 1]
    \tiny
    \textbf{Candidate 1}
    \par\vspace{2mm}
    Text:
    Blow... watching them pick, process, paste, powder, pack, plane, and play Can't You Hear Me Knockin' through the whole thing was pretty epic.
    \par
    \par
    Text:
    If the sun is up, we're up, and feed me, after that you can go back to sleep. Hard food isn't food. Hard food is emergency backup food for if you leave me home too long without refilling my soft food dish.
    \par
    Text:
    The Members Only jacket is choice..... I had a black one myself.
    \par
    Text:
    I'm 43 years old and find this to be true for me as well.
    \par\vspace{2mm}
    {\bf Explanation of the Style}
    The writing style of these authors is characterized by a dynamic interplay of informal language and vivid emotional expression, rendering their work inherently approachable and deeply personal. [...]. Interspersed with enthusiastic interjections such as "Wow" and "Exquisite!" along with stark emotional outbursts like "FUCK THEM!", the authors convey their impassioned perspectives with raw authenticity. [...] This is further amplified by their deft use of direct address, posing questions and thoughts like "Is that better?" to forge an intimate connection, sparking personal reflection or humor, leaving a lasting impact.

\tcbitem[title=Candidate 2]
    \tiny
    \textbf{Candidate 2}
    \par\vspace{2mm}
    Text:
My mantra was to always follow the old adage of never reading the comments on articles especially if it involves race. Then, [...]. You're shining something on a social issue which needs to be addressed. As they say, "light is the best disinfectant" and "truth does not fear investigation."

Keep going.

As an Asian guy, the shows I found the most interesting growing up was all about urban black culture. Boyz N The Hood. Juice. Friday. Menace II Society. Dead Presidents. New Jack City. Do The Right Thing. Baby Boy.

Text:
It's assumed that anything worth knowing is already known by white people. If they aren't aware of something specific, it's most likely useless.

Or someone cheated by knowing it first.

\par\vspace{2mm}
{\bf Explanation of the style}
The authors' writing style is characterized by a friendly, conversational tone that invites readers into an almost intimate dialogue, blurring the lines between spoken and written word. Through the use of informal language peppered with colloquialisms, slang, and relaxed punctuation, they cultivate a warm, approachable voice that feels personal and engaging. [...]. Ultimately, their prose is imbued with a spectrum of emotions — from humor to empathy — which resonate loudly with readers, rendering their writing not just a medium of information, but a vivid tapestry of experiences and heartfelt expression.

\tcbitem[
    title=Candidate 3 (Predicted Author),
    colframe=candidateGreen 
]
    \tiny
    \textbf{\textcolor{candidateGreen}{Candidate 3}}
    
    \textbf{\textcolor{candidateGreen}{(Predicted Author)}}
    \par\vspace{2mm}

    Text:
Awesome, thank you! You get the Most Responsive Dev award™ :)

Text:
"TSF Launcher 3D Shell" is apparently closer to "Shell Launcher" than "Shell Launcher". Nobody does search better than Google /s
\par
Text:
Why rely on implementation details when there's a trivial way to do what you want in a portable manner?
\par
Text:
Nope, none at all.

I don't actually know that, but it seems like the kind of thing that won't happen unless you're Sundar Pichai.
\par
Text:
You should still use OrderdDict anyway - this is only an implementation detail!
\par\vspace{2mm}
{\bf Explanation of the style}
The writing style paints a portrait of intimacy and engagement, marked by an informal and conversational tone that invites readers into a dialogue rather than a monologue. [...]. They frequently link out to external sources, weaving in a tapestry of multimedia elements to add depth and context to their narratives. This openness extends to their use of parenthetical comments, strategically inserted to inject humor, add depth, or clarify their points without breaking the rhythm of the prose. [...], which infuse each piece with a light-heartedness that makes even the most complex subjects relatable and enchanting for readers.

    \par\vspace{2mm}
    
\end{tcbitemize}

\end{tcolorbox}

\caption{Example baseline explanations for the model's prediction on one of the AA tasks}
\label{fig:baseline-explanations}
\end{figure*}

\begin{figure*}
    \centering
    \includegraphics[]{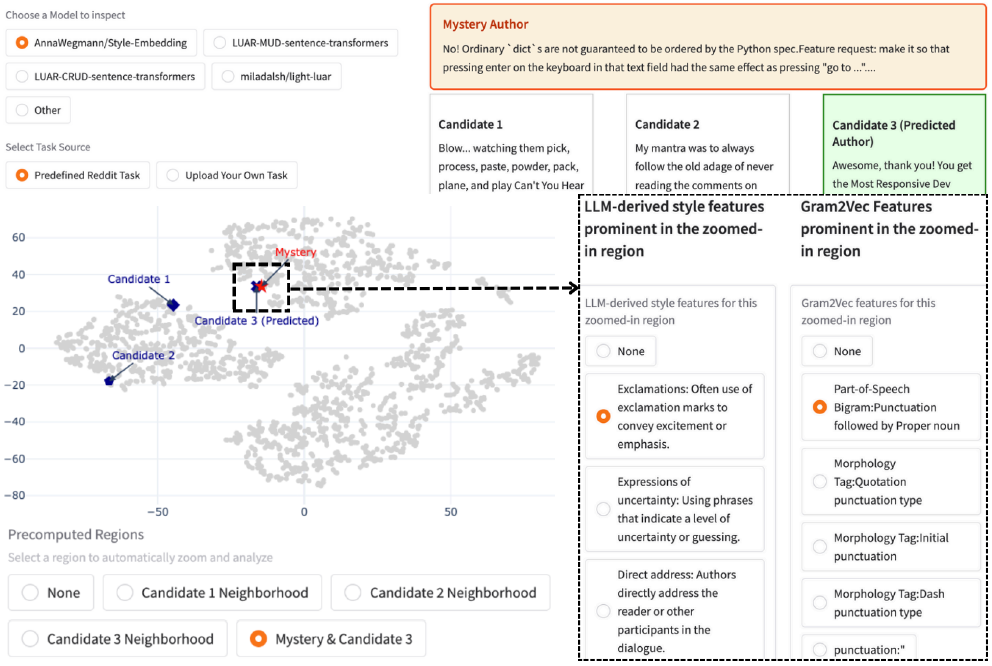}
    \caption{Our tool allows users to visualize the latent space of the AA model and inspect a relevant subregion to see its common writing style features.}
    \label{fig:main-approach}
\end{figure*}

\begin{figure*}
    \centering
    \includegraphics[]{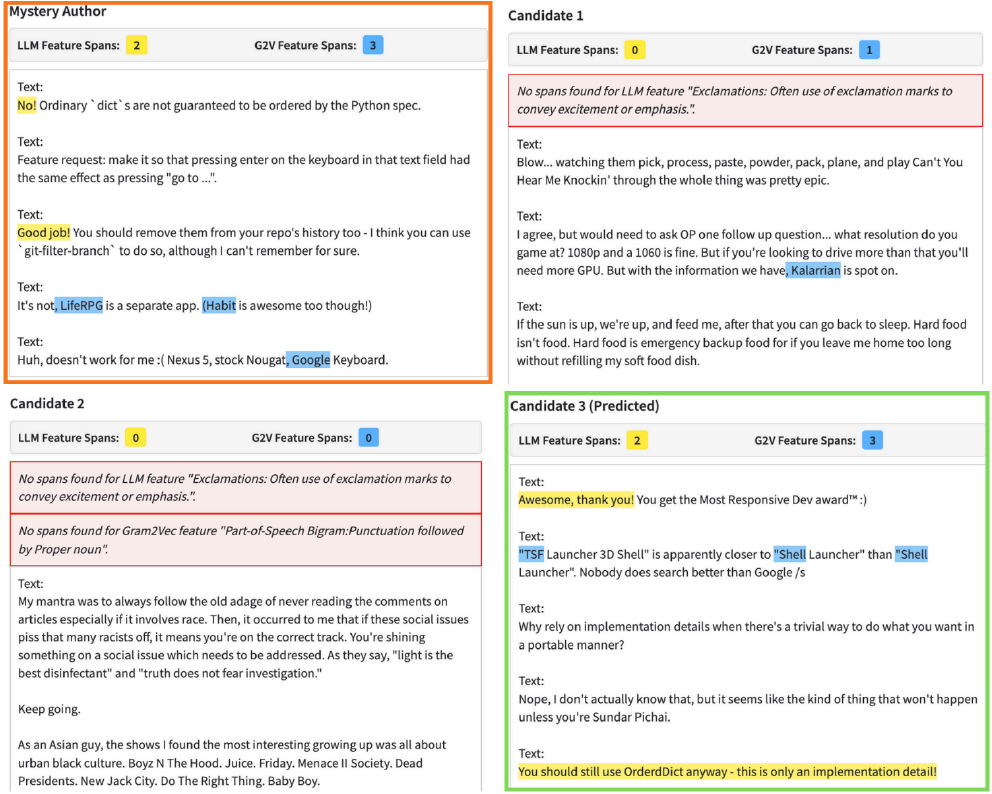}
    \caption{Our tool allows users to show example spans from the text of the task authors for each of the features}
    \label{fig:main-approach}
\end{figure*}
\label{sec:appendix}


\end{document}